\documentclass{article}

\usepackage{tikz}
\usetikzlibrary{shapes,decorations,arrows,calc,arrows.meta,fit,positioning}
\tikzset{
    -Latex,auto,node distance =1 cm and 1 cm,semithick,
    state/.style ={ellipse, draw, minimum width = 0.7 cm},
    point/.style = {circle, draw, inner sep=0.04cm,fill,node contents={}},
    bidirected/.style={Latex-Latex,dashed},
    el/.style = {inner sep=2pt, align=left, sloped}
}

\usepackage{microtype}
\usepackage{graphicx}
\usepackage{subfigure}
\usepackage{booktabs} 

\usepackage{hyperref}

\usepackage[accepted]{icml2023}

\usepackage{amsmath}
\usepackage{amssymb}
\usepackage{mathtools}
\usepackage{bbm}
\usepackage{amsthm}

\usepackage{amsmath,amsthm,amssymb}
\newcommand{\EE}[2][]{\mathbb{E}_{#1}\left[#2\right]}

\newcommand{\cX}{\mathcal X}
\newcommand{\cC}{\mathcal C}
\newcommand{\EEE}{\mathbb{E}}
\newtheorem{assu}{Assumption}
\newtheorem{prop}{Proposition}
\newtheorem{thm}{Theorem}
\newtheorem{rema}{Remark}
\newtheorem{alg}{Algorithm}

\usepackage[capitalize,noabbrev]{cleveref}

\theoremstyle{plain}

\theoremstyle{definition}

\theoremstyle{remark}

\usepackage[textsize=tiny]{todonotes}

\icmltitlerunning{Proximal Causal Learning of Conditional Average Treatment Effects}

\begin{document}

\twocolumn[
\icmltitle{Proximal Causal Learning of Conditional Average Treatment Effects}



\icmlsetsymbol{equal}{*}

\begin{icmlauthorlist}
\icmlauthor{Erik Sverdrup}{yyy}
\icmlauthor{Yifan Cui}{aaa}
\end{icmlauthorlist}

\icmlaffiliation{yyy}{Graduate School of Business, Stanford University, Stanford, USA.}
\icmlaffiliation{aaa}{Center for Data Science, Zhejiang University, Hangzhou, China}

\icmlcorrespondingauthor{Erik Sverdrup}{erikcs@stanford.edu}
\icmlcorrespondingauthor{Yifan Cui}{cuiyf@zju.edu.cn}

\icmlkeywords{Causal Inference, Double Robustness, Proximal Causal Learning, Treatment Heterogeneity, Unmeasured Confounders}

\vskip 0.3in
]



\printAffiliationsAndNotice{}  

\begin{abstract}
Efficiently and flexibly estimating treatment effect heterogeneity is an important task in a wide variety of settings ranging from medicine to marketing, and there are a considerable number of promising conditional average treatment effect estimators currently available. These, however, typically rely on the assumption that the measured covariates are enough to justify conditional exchangeability. We propose the \emph{P-learner}, motivated by the \emph{R-} and \emph{DR-learner}, a tailored two-stage loss function for learning heterogeneous treatment effects in settings where exchangeability given observed covariates is an implausible assumption, and we wish to rely on proxy variables for causal inference. Our proposed estimator can be implemented by off-the-shelf loss-minimizing machine learning methods, which in the case of kernel regression satisfies an oracle bound on the estimated error as long as the nuisance components are estimated reasonably well.
\end{abstract}

\section{Introduction}
The conditional average treatment effect (CATE) measures the net benefit, such as the decrease in blood pressure, a certain subset of a population experiences by being assigned a certain intervention, such as a drug. Obtaining accurate estimates of CATEs is important in order to understand for example which parts of a population should be assigned a treatment, if any. A large body of work has made tremendous advances in designing flexible estimators for the CATE. Examples include \citet{hill2011bayesian, alaa2017bayesian, hahn} for Bayesian approaches, \citet{atheyimbens, wagerathey} for tree-based methods, \citet{johansson2016learning, shalit2017estimating, yoon2018ganite, shi2019adapting} for adopting neural networks, and \citet{kunzel, nie2020quasi} for combinations thereof.

To identify causal effects, the aforementioned approaches operate under the exchangeability assumption, i.e., the assertion that conditional on observed covariates, the treatment assignment is as good as random. We propose a CATE estimator, which using the framework of \citet{tchetgen2020introduction}, allows one to estimate causal effects in settings where conditional exchangeability \emph{fails}, but one has measured a set of sufficient \emph{proxy} variables. Our practical approach is motivated by the generic Neyman-orthogonal \citep{chernozhukov2018double} loss function from \citet{nie2020quasi} and \citet{kennedy2020towards} that decouples nuisance estimation and CATE estimation into two stages that can be estimated (and tuned with cross-validation) by flexible loss-minimizing machine learning tools, where the latter stage is to first order less sensitive to estimation error arising from the first stage. Our contribution is to extend this flexible CATE estimation strategy to the proximal causal inference framework \citep{tchetgen2020introduction}. The proposed loss relies on doubly robust scores \citep{robins1994estimation,rotnitzky1998semiparametric,scharfstein1999adjusting,chernozhukov2018double,cui2020semiparametric} which can also be re-purposed to enable semi-parametric efficient estimation and inference on lower-dimensional summaries of the CATEs, such as best linear projections \citep{semenova2021debiased}, or rank-weighted average treatment effects \citep{yadlowsky2021evaluating}.

\subsection{Proxy Variables and Unmeasured Confounding}
Conditional exchangeability (often also referred to as unconfoundedness \citep{imbens2015causal}), is a crucial identifying assumption that underlies many popular methodologies for estimating causal effects from observational data, including most CATE estimators. Loosely stated, it requires the investigator to have collected a sufficient set of covariates, such that controlling for these, the treatment assignment is as good as random. Given some additional regularity assumptions, this allows the investigator to estimate a difference in potential outcomes, without having access to a randomized control trial \citep{imbens2015causal}. Naturally, the quality of the causal estimates hinges on whether the collected covariates sufficiently account for confounding, and a large body of work, going back to for example \citet{rosenbaum1983assessing} has pioneered measures for assessing the sensitivity of a causal estimate to this assumption. 

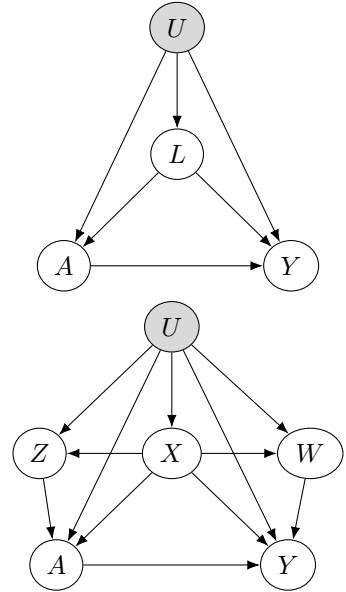
\begin{figure}[ht!]
\centering
\begin{tabular}{c}
\begin{tikzpicture}
    \node[state] (l) at (0,0) {$L$};
    \node[state] (y) [below right=of l] {$Y$};
    \node[state] (a) [below left=of l] {$A$};
    \node[state,fill=black!15] (u) [above=of l] {$U$};
    \path (l) edge (y);
    \path (l) edge (a);
    \path (u) edge (a);
    \path (u) edge (l);
    \path (u) edge (y);
    \path (a) edge (y);
\end{tikzpicture} \\

\begin{tikzpicture}
    \node[state] (x) at (0,0) {$X$};
    \node[state] (y) [below right=of x] {$Y$};
    \node[state] (w) [right=of x] {$W$};
    \node[state] (a) [below left=of x] {$A$};
    \node[state] (z) [left=of x] {$Z$};
    \node[state,fill=black!15] (u) [above=of x] {$U$};
    \path (x) edge (y);
    \path (x) edge (z);
    \path (x) edge (a);
    \path (x) edge (w);
    \path (u) edge (z);
    \path (u) edge (a);
    \path (u) edge (x);
    \path (u) edge (y);
    \path (u) edge (w);
    \path (z) edge (a);
    \path (a) edge (y);
    \path (w) edge (y);
\end{tikzpicture}
\end{tabular}
\caption{Top: an illustration of a violation of conditional exchangeability due to the presence of the unmeasured confounder $U$ that affects both the treatment $A$ and the outcome $Y$, even given observed covariates $L$. Bottom: if one is able to partition $L$ into $X$ (common causes of $A$ and $Y$), $Z$ (proxies for confounders that affect $A$), and $W$ (proxies for confounders that affect $Y$), then these may be used to back out the implied bias caused by $U$. Note that exchangeability need not hold conditional on observed covariates.}
\label{fig:figDAG}
\end{figure}

The proximal causal inference framework \citep{tchetgen2020introduction} departs from this classical approach by instead asking: even with the presence of unmeasured confounding, are the alternative and realistic assumptions that can be made in order to estimate causal effects? The answer is yes: if the investigator has access to auxiliary variables that satisfy certain assumptions. These auxiliary variables are so-called \emph{proxy} variables that augment the set of controls with additional variables that are either treatment-inducing or outcome-inducing. Consider a simple example where we are interested in the causal effect of the treatment $A$ on the outcome $Y$ and have collected a set of covariates $L$ that is related to both the treatment and outcome. Unfortunately, due to the presence of an unmeasured confounder $U$, conditional exchangeability fails. Figure \ref{fig:figDAG} (top) shows this scenario as a causal DAG \citep{pearl2009causality}. The proximal causal learning framework relaxes the conditional exchangeability assumption and instead operates under the premise that the variables $L$ can be partitioned into three specific groups: common causes of the treatment and outcomes ($X$), treatment-inducing confounding proxies ($Z$), and outcome-inducing confounding proxies ($W$). Figure \ref{fig:figDAG} (bottom) show the covariates $L$ partitioned into these three groups and suggest that with reasonable assumptions on the interdependencies between treatment, outcomes, confounders, and proxies, one may still learn causal effects. The intuition is that we may use this structure to back out the net effect of the unobserved confounder through its relation with $Z$ and $W$, then remove this confounding bias to arrive at the effect of $A$ on $Y$. As prudently pointed out in \citet{tchetgen2020introduction}, many observational datasets exhibit a certain structure whereby the data collected was not measured with the precise intent of quantifying a certain source of confounding. Rather, depending on the question the investigator is trying to answer, these collected variables serve as noisy measures of confounding. The promise of proximal causal learning is that with the right assumptions and structure, one may leverage a subset of these noisy variables as ``proxies'' which serve the purpose of backing out the net effect of the confounder $U$.

\subsection{Previous Work}
There is a fast-growing literature on causal inference methods that leverage proxy variables to mitigate confounding \citep{lipsitch2010negative, kuroki2014measurement,deaner2018proxy}. On a high level, our work leverages the generalizations set forth by \citet{tchetgen2020introduction} who cast proximal causal learning in the potential outcomes framework and \citet{miao2018identifying,cui2020semiparametric} who provide nonparametric identification results for average treatment effects.
\citet{dukes2021proximal, shi2021theory,shpitser2021proximal,ying2021proximal,ying2022proximal} consider identification and estimation for other causal quantities. 

A nascent body of work is developing new methods utilizing this framework to answer important questions. \citet{singh2020kernel} develop kernel methods for nonparametric estimation of for example dose-response curves under unmeasured confounding, \citet{li2022double} use proximal causal learning to estimate vaccine effectiveness, \citet{qi2023proximal, shen2022optimal} develop individualized treatment allocation rules under unmeasured confounding, and \citet{imbens2021controlling} propose novel methods for panel data with proxies. Besides causal inference, the core idea of the proximal causal inference framework has also been adopted in survival analysis \citep{ying2022proximalb} to address dependent censoring and reinforcement learning \citep{shi2021minimax,bennett2021proximal} to tackle a partially observable Markov decision process.

Our practical approach for CATE estimation draws inspiration from \citet{nie2020quasi} who cast the problem as a generic two-step loss minimization (the \emph{R-learner}, named so to commemorate the work of \citet{robinson1988} for the statistical foundation of the approach) that can be implemented by off-the-shelf machine learning methods. The benefit of this decoupling is that it clearly separates the statistical tasks of estimating nuisance components, and that of estimating treatment effects, which can be implemented and optimized (by standard cross-validation) through different learners (\citet{wu2022integrative} extend the \emph{R-learner} to handle data combination from observational and experimental trial data).

The final step of our approach takes the form of a pseudo outcome regression, where transformed outcomes are regressed on covariates, and this approach dates back to \citet{van2006statistical, luedtke2016super} who under traditional unconfoundedness suggests it as a method for estimating CATEs, but without explicit error guarantees. \citet{kennedy2020towards} and \citet{foster2019orthogonal} give error guarantees under general assumptions on the nuisance components (when estimated using sample splitting) and derive desirable properties for this approach to CATE estimation (in Section \ref{sec:method} we highlight the connection between \citet{kennedy2020towards}'s proposed \emph{DR-learner} under unconfoundedness with the proximal \emph{P-learner}, and how our approach can be seen as a generalization of existing methods for CATE estimation under unconfoundedness to proxies). Finally, classical approaches to causal estimation, when unconfoundedness fails, is to rely on instruments, and \citet{syrgkanis2019machine} develop a powerful loss-based method to estimate treatment effects conditional on the units complying with the instrument.

\section{CATE Estimation under Unmeasured Confounding}\label{sec:CATE}
\subsection{Setup}\label{sec:setup}
We operate under the potential outcomes framework and posit the existence of potential outcomes $Y_i(1), Y_i(0)$ corresponding to binary treatment assignment $A_i=\{0, 1\}$. We have access to a collection of covariates $X_i$ that are potential common causes of both the treatment and the outcome and are interested in measuring the conditional average treatment effect (CATE) defined as $\tau^*(X_i) = \EE{Y_i(1) - Y_i(0) | X_i = x}$. We assume that conditioning on the covariates $X_i$ is not sufficient to guarantee conditional exchangeability, due to the presence of latent unmeasured confounders $U_i$. To identify $\tau^*(x)$ we rely on the existence of proxy variables $Z_i$ and $W_i$ where $Z_i$ are treatment inducing and $W_i$ outcome inducing, which satisfies

\begin{assu}\label{assu:iden1}
$Y_i \perp Z_i | U_i , X_i, A_i$.
\end{assu}
\begin{assu}\label{assu:iden2}
$W_i \perp (Z_i, A_i) | U_i, X_i$.
\end{assu}
\begin{assu}\label{assu:iden3}
$(Y_i(1), Y_i(0)) \perp A_i | U_i, X_i$.
\end{assu}
\begin{assu}\label{assu:iden4}
For any square-integrable function g and for any a, x, $\EE{g(U_i) | Z_i, A_i = a, X_i = x}=0$ almost surely if and only if $g(U_i)=0$ almost surely.
\end{assu}
\begin{assu}\label{assu:iden5}
For any square-integrable function g and for any a, x, $\EE{g(U_i) | W_i, A_i = a, X_i = x}=0$ almost surely if and only if $g(U_i)=0$ almost surely.
\end{assu}

The last two assumptions are completeness conditions that essentially ensure that the proxy variables carry sufficient information about the confounders $U_i$.  Given consistency assumptions on potential outcomes, positivity assumptions on the treatment assignment $A_i$, and some technical regularity conditions \citep{miao2018identifying,cui2020semiparametric}, this setup allows for identification of causal effects through assuming the existence of the following integral equations:
\begin{equation}\label{eq:hbridge}
    \EE{Y_i |Z_i, A_i, X_i} = \int h^*(w, A_i, X_i)dF(w|Z_i, A_i, X_i),
\end{equation}
and
\begin{equation}\label{eq:qbridge}
    \EE{q^*(Z_i, a, X_i)|W_i, A_i = a, X_i} = \frac{1}{f(A_i = a|W_i, X_i)},
\end{equation}
almost surely.
In the proximal causal inference literature, Equations~\eqref{eq:hbridge} and \eqref{eq:qbridge} are referred to as bridge functions and characterize a type of inverse problem (known as the Fredholm integral equation) that allows for the identification of counterfactual means with the presence of unmeasured confounders $U_i$. We defer a discussion of obtaining estimates of $h^*$ and $q^*$ to Section~\ref{sec:estimatenuis}.

\subsection{A Loss Function for Proximal CATEs}\label{sec:method}
To estimate an \emph{average} treatment effect $\EE{Y_i(1) - Y_i(0)}$ under proximal causal learning, motivated by the semiparametric efficient influence function given in \citet{cui2020semiparametric}, we consider the following doubly robust score, 
\begin{equation}\label{eq:psi}
\begin{split}
  &\Gamma_i = (-1)^{1-A_i}q^*(Z_i, A_i, X_i)\left(Y_i - h^*(W_i, A_i, X_i)\right)\\
  &+ h^*(W_i, 1, X_i) - h^*(W_i, 0, X_i).
\end{split}
\end{equation}
The corresponding semiparametric efficient ATE estimate is the sample average of $\Gamma_i$. The key insight is to recognize that to efficiently estimate a \emph{conditional} ATE (CATE), it suffices to learn the mapping from covariates $X$ to pseudo-outcome $\Gamma$. This motivates the following procedure:

\begin{alg}{(P-learner)}
\end{alg}
\begin{description}
    \item[Step 1.] Split the data, $i=1 \ldots n$, into $C$ evenly sized folds. Estimate $h(w, a, x)$ and $q(z, a, x)$ with cross-fitting over the $C$ folds, using tuning as appropriate.

    \item[Step 2.] Form the scores \eqref{eq:psi} using cross-fit plug-in estimates of nuisance components $\hat h^{(-c(i))}(z, a, x)$ and $\hat q^{(-c(i))}(z, a, x)$, where the notation $c(\cdot)$ maps from sample to fold and $(-c(i))$ indicates predictions made without using the $i$-th sample for training. We estimate treatment effects by minimizing the following empirical loss
\begin{equation}\label{eq:plearner}
    \hat \tau(\cdot) = \arg\min_\tau \left[ \hat L_n(\tau(\cdot)) \right],
\end{equation}
where
\begin{equation}\label{eq:plearnereq}
    \hat L_n(\tau(\cdot))= \frac{1}{n} \sum_{i=1}^n \left(\hat \Gamma^{(-c(i))}_i - \tau(X_i) \right)^2
\end{equation}
and $\hat \Gamma^{(-c(i))}_i$ are cross-fit estimates of the scores \eqref{eq:psi}, i.e.,
\begin{align*}
\hat \Gamma^{(-c(i))}_i &= (-1)^{1-A_i}\hat q^{(-c(i))}(Z_i, A_i, X_i)\\&
\times \left(Y_i - \hat h^{(-c(i))}(W_i, A_i, X_i)\right)
  \\& + \hat h^{(-c(i))}(W_i, 1, X_i) - \hat h^{(-c(i))}(W_i, 0, X_i).
\end{align*}
\end{description}

The final stage model $\EE{\hat \Gamma^{(-c(i))}_i | X_i = x}$ is purely a predictive problem that can leverage flexible non-parametric learners ranging from random forests to neural networks, etc (or combinations in the form of stacking), to simpler parametric models.  As shown in Section \ref{sec:theory}, this approach can deliver accurate estimates of $\tau^*(\cdot)$ even when the nuisance components are subject to estimation error. 

An interesting conceptual connection is that under the traditional unconfoundedness assumption, Equation \eqref{eq:psi} reduces to the celebrated Augmented Inverse-Probability Weighted (AIPW) score of \citet{robins1994estimation, robins1995semiparametric}, which forms the basis for the \emph{DR-learner} proposed by \citet{kennedy2020towards} where the unconfounded efficient influence function for the ATE is used to learn a CATE function via a similar procedure. 

\begin{rema}
The empirical loss \eqref{eq:plearnereq} can be used for learning other estimands of interest. For example, for the conditional average treatment effect on the treated $\mu^*(x) = \EE{Y_i (1) - Y_i (0)|A_i = 1, X_i=x}$, suppose there exists $h^*$ and $q^*$ satisfying 
\begin{equation*}
    \EE{Y_i |Z_i, A_i = 0, X_i} = \int h^*(w, X_i)dF(w|Z_i, A_i = 0, X_i),
\end{equation*}
and
\begin{equation*}
    \EE{q^*(Z_i, X_i)|W_i, A_i = 0, X_i} = \frac{f(A_i = 1|W_i, X_i)}{f(A_i = 0|W_i, X_i)}.
\end{equation*}
Motivated by \citet{cui2020semiparametric}, we define the following loss function,
\begin{align*}
 & \hat L^{CATOT}_n(\mu(\cdot))= \frac{1}{n} \sum_{i=1}^n [
 A_iY_i \\&- (1 - A_i)\hat q^{(-c(i))}(Z_i, X_i)[Y_i - \hat h^{(-c(i))}(W_i, X_i)]\\ &- A_i[\hat h^{(-c(i))}(W_i, X_i) + \mu(X_i)]]^2.
\end{align*}
The rest of the learning procedure follows Algorithm 1.
\end{rema}

\subsection{Doubly Robust Scores and Semiparametric Inference}
While the aforementioned procedure for learning $\tau(\cdot)$ satisfies desirable oracle properties, it is still a challenging statistical task to conduct pointwise inference on the estimated treatment effects (see for example \citet{armstrong2018optimal} for fundamental limits to uncertainty quantification for a single point estimated nonparametrically). For this reason, it is often advisable to delegate uncertainty quantification to lower-dimensional summaries of the $\tau(\cdot)$ function. This is the approach advocated by \citet{chernozhukov2018double} and can be attained by the doubly robust score construction in Equation \eqref{eq:psi}. As a concrete example, if we in \emph{Step 2} fit a linear parametric model, we recover the best linear projection of \citet{semenova2021debiased, chernozhukov2018generic} that delivers standard errors with nominal coverage (demonstrated as an example in Section \ref{sec:application}). Moreover, for assessing if the estimated $\tau(\cdot)$ function actually manages to stratify the population into groups that respond differently to treatment, the same score construction \eqref{eq:psi} can be used to construct the recently proposed RATE metric of \citet{yadlowsky2021evaluating}. We present an example of this approach in Section \ref{sec:applicationRATE}.

\subsection{Estimating Nuisance Components}\label{sec:estimatenuis}
As mentioned in Section \ref{sec:setup}, two crucial ingredients for proximal causal learning are the bridge functions $h^*$ and $q^*$. Equations~\eqref{eq:hbridge} and \eqref{eq:qbridge} define challenging inverse problems known as Fredholm integral equations of the first kind. Timely and pioneering work by \citet{dikkala2020minimax} gives an empirical strategy for estimating these quantities by regularized minimax estimation, which \citet{ghassami2022minimax} draws on to propose a flexible kernel machine learning estimator of $h^*$ and $q^*$ (this is also the approach used by \citet{kallus2021causal,mastouri2021proximal,qi2023proximal}).

For the purpose of simulated and practical illustrations of the \emph{P-learner}, we rely on the flexible kernel estimator of \citet{kallus2021causal, ghassami2022minimax}, which can be implemented by off-the-shelf software using Gaussian kernels and tuned with cross-validation.
In particular, we consider the following min-max optimization problem,
\begin{align*}
 &\min_{h \in \mathcal{H}} \max_{r \in \mathcal{R}}\mathbb{P}_n[(\mathbb{I}\{A = a\}Y -\mathbb{I}\{A = a\}h(W,a,X))r(Z,X)\\&~~~~~~~~~~~~~~~~~~~~~~~   -r^2(Z,X)] - \lambda_r||r||^2_{\mathcal{R}} +\lambda_h||h||^2_{\mathcal{H}},\\
&\min_{q \in \mathcal{Q}} \max_{s \in \mathcal{S}}\mathbb{P}_n[(1-\mathbb{I}\{A=a\}q(Z,a,X))s(W,X)\\&~~~~~~~~~~~~~~~~~~~~~~~  -s^2(W,X)] - \lambda_s||s||^2_{\mathcal{S}} +\lambda_q||q||^2_{\mathcal{Q}},
\end{align*}
where $\mathcal{R}, \mathcal{H}, \mathcal{S}$, and $\mathcal{Q}$ are critic classes with norms represented by $||\cdot||_{\mathcal{R}}, ||\cdot||_{\mathcal{H}}, ||\cdot||_{\mathcal{S}}$, and $||\cdot|| _{\mathcal{Q}}$,  $\mathbb{P}_{n}$ denotes the sample average with respect to a training sample, and $\lambda_Q^h, \lambda_H^h, \lambda_H^q, \lambda_Q^q$ are tuning parameters. This problem has a closed-form solution given by Propositions~9 and 10 in \citet{dikkala2020minimax}. 

\section{Oracle Bound for \emph{P-learner}}\label{sec:theory}

Our theory focuses on a \emph{P-learner} based on penalized kernel regression. Regularized kernel learning has been thoroughly studied in the learning literature \cite{bartlett2006empirical,steinwart2008support,mendelson2010regularization} and is also used in the theoretical analysis of the R-learner \cite{nie2020quasi}. Following \citet{nie2020quasi}, we study $||\cdot||_{\cC}$-penalized kernel regression, where $\cC$ is a reproducing kernel Hilbert space (RKHS) with a continuous, positive semi-definite kernel function.

Our main goal is to establish error bounds for \emph{P-learner} that only depends on the complexity of $\tau^*(\cdot)$, and that match the error bounds we could achieve if we knew $h^*$ and $q^*$ a priori. 
We study the following cross-fitted estimator and its oracle analog
\begin{align*}
\hat \tau (\cdot) =& \arg\min_\tau \Big[ \frac{1}{n} \sum_{i=1}^n \left(\hat \Gamma^{(-c(i))}_i - \tau(X_i) \right)^2 \\&+ \lambda_n(||\tau||_{\cC}) : ||\tau||_\infty\leq 2M\Big],\\
\tilde \tau(\cdot) = & \arg\min_\tau \Big[ \frac{1}{n} \sum_{i=1}^n \left(\Gamma_i - \tau(X_i) \right)^2\\& + \lambda_n(||\tau||_{\cC}) : ||\tau||_\infty\leq 2M\Big],
\end{align*}
respectively, where $\lambda_n(||\tau||_{\cC})$ is a properly chosen penalty. 
Similar to the estimated loss given in~\eqref{eq:plearnereq}, we define population and oracle losses 
\begin{align*}
    L(\tau(\cdot))&= \EE{\left(\Gamma_i - \tau(X_i) \right)^2},\\
        \tilde L_n(\tau(\cdot))&= \frac{1}{n} \sum_{i=1}^n \left(\Gamma_i - \tau(X_i) \right)^2,
\end{align*}
respectively. We are interested in the regret bound $R(\tau) = L(\tau(\cdot)) - L(\tau^*(\cdot))$ for our \emph{P-learner} $\hat \tau(\cdot)$.

Let 
\begin{align*}
\cC_\alpha = \{\tau: ||\tau||_{\cC} \leq \alpha, ||\tau||_{\infty} \leq 2M\}
\end{align*}
denote a radius-$\alpha$ ball of $\cC$ capped by $2M$. 
We denote
$\tau_\alpha^* = \arg\min \{L(\tau) : \tau \in \cC_\alpha\}$
as the best approximation to $\tau^*$ within the working class $\cC_\alpha$. 

In addition, we also define the population, estimated, and oracle $\alpha$-regret functions
\begin{align*}
R(\tau, \alpha) &= L(\tau) - L(\tau_\alpha^*),\\
\hat R_n(\tau, \alpha) &= \hat L_n(\tau) - \hat L_n(\tau_\alpha^*),\\
\tilde R_n(\tau, \alpha) &= \tilde L_n(\tau) - \tilde L_n(\tau_\alpha^*),
\end{align*}
respectively. 
Then we are ready to state the following proposition. 

\begin{prop}\label{prop}
Suppose Assumptions~\ref{assu:iden1}-\ref{assu:iden5} and Equations~\eqref{eq:hbridge}-\eqref{eq:qbridge} hold. Further assume Assumptions~\ref{ass:kernel}-\ref{assu:bound} given in the Appendix hold, then we have 
\begin{align*}
\hat R(\tau,\alpha) - \tilde R(\tau,\alpha)
\leq O_P \Bigg(  \alpha^{p}  n^{-1/2} R(\tau,\alpha)^{\frac{1-p}{2}} + \\
\alpha^p \frac{\log(n)}{n^{3/4}}  R(\tau,\alpha)^{\frac{1-p}{2}}+  \frac{\alpha^p}{n^{3/4}} \sqrt{\log(  \frac{\alpha n^{1/(1-p)}}{R(\tau,\alpha)}) } R(\tau,\alpha)^{\frac{1-p}{2}}\\+ \frac{\alpha^p}{n} \log(  \frac{\alpha n^{1/(1-p)}}{R(\tau,\alpha)}) R(\tau,\alpha)^{\frac{1-p}{2}}+  \frac{\alpha^p}{n^{5/4}}R(\tau,\alpha)^{\frac{1-p}{2}}  \Bigg),
\end{align*}
where $p$ is defined in Assumption~\ref{ass:kernel} given in the Appendix.
\end{prop}

The proof of Proposition~\ref{prop} is given in the Appendix. 
This key result provides the excess error bound for regret $\hat R(\tau,\alpha)$ using the cross-fitted learner over the regret bound $\tilde R(\tau,\alpha)$ using the oracle learner.
By leveraging the above proposition, we have the following theorem.

\begin{thm}\label{thm}
Suppose the conditions of Proposition~\ref{prop} hold. With a properly chosen penalty $\lambda_n(||\tau||)$, $\hat \tau(\cdot)$ satisfies the same regret bound $\tilde \tau(\cdot)$, that is,  $R (\hat \tau ) =R (\tilde \tau ) = O_P(n^{-(1-2\beta)/[p+(1-2\beta)]})$, where $\beta$ is defined in Assumption~\ref{ass:operator} given in the Appendix.
\end{thm}

Note that the \emph{P-learner} objective is the following regression: $\hat \tau(\cdot) =
\arg\min_{\tau\in \cC_\alpha} \frac{1}{n} \sum_{i=1}^n \left(\hat \Gamma^{(-c(i))}_i - \tau(X_i) \right)^2$. The proof of Theorem~\ref{thm} is essentially the same as Theorem~3 of \citet{nie2020quasi} and we omit it here. 
Theorem~\ref{thm} implies that with penalized kernel regression, the cross-fitted \emph{P-learner} can achieve a similar performance as the oracle learner that knows both confounding bridge functions a priori.

\section{A Motivating Example}\label{sec:illustration}
To illustrate the promise of the \emph{P-learner} we consider a simple motivating example that highlights some salient features (to the best of our knowledge, we are not aware of other current proposals for proximal CATE estimation, making traditional benchmark comparisons challenging). We design a proximal data generating mechanism using the setup from \citet{cui2020semiparametric} where we incorporate treatment heterogeneity using the moderately complex CATE function $\tau^*(X) = \exp(X_{(1)}) - 3X_{(2)}$ used in \citet{shen2022optimal} to learn proximal treatment regimes and add three additional irrelevant normally distributed covariates $X$ (the complete setup is described in Appendix \ref{sec:DGP}). In the left-most plot in Figure \ref{fig:cfVSlasso} we train a \emph{Causal Forest} \citep{athey2019generalized}, a popular method for estimating CATEs under conditional exchangeability on data with $n_{train}=4000$ samples and predict the estimated CATEs on a test set with $n_{test}=2000$. Since unconfoundedness fails, the point estimates are considerably biased, as seen by estimated CATEs falling above the 45-degree line shown in red.
\begin{figure}[ht!]
\centering
    \includegraphics[scale=0.6]{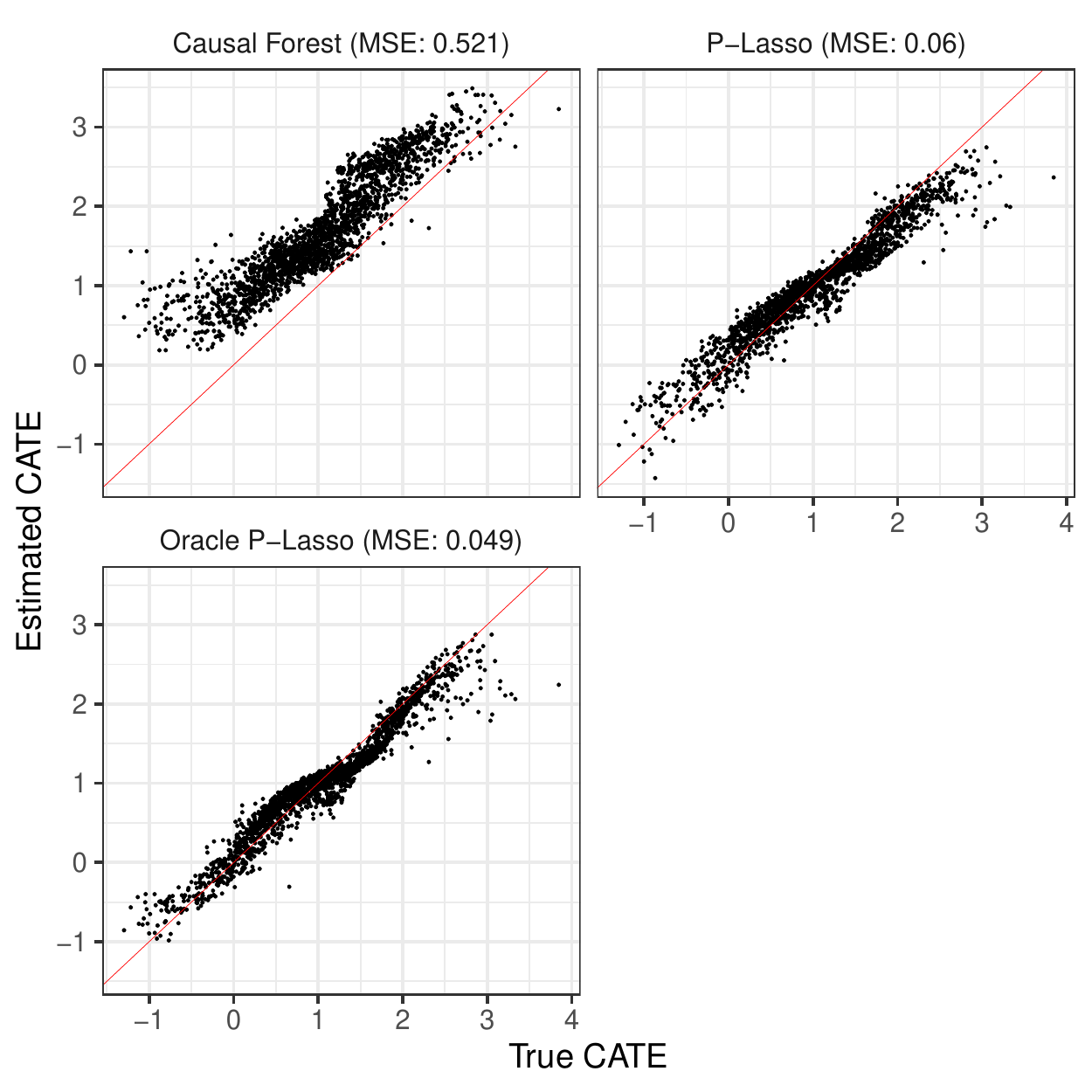}
\caption{Top left: a \emph{Causal Forest} \citep{athey2019generalized} fit on simulated training data ($n_{train}=4000$ and 5 covariates) where unconfoundedness fails, is used to predict estimated CATEs on a test set ($n_{test}=2000$). Top right: estimated and true CATEs using the same training and test data with a \emph{P-learner} using cross-validated Lasso \citep{friedman2010regularization} and nuisance components $h$ and $q$ estimated with kernels \citep{ghassami2022minimax}. Bottom left: the same \emph{P-learner} fit using oracle $h$ and $q$. Mean square error (MSE) is defined as $\frac{1}{n}\sum_{i=1}^{n_{test}}(\hat \tau(X_i) - \tau^*(X_i))^2$.}
\label{fig:cfVSlasso}
\end{figure}

\begin{figure}[ht!]
\centering
    \includegraphics[scale=0.6]{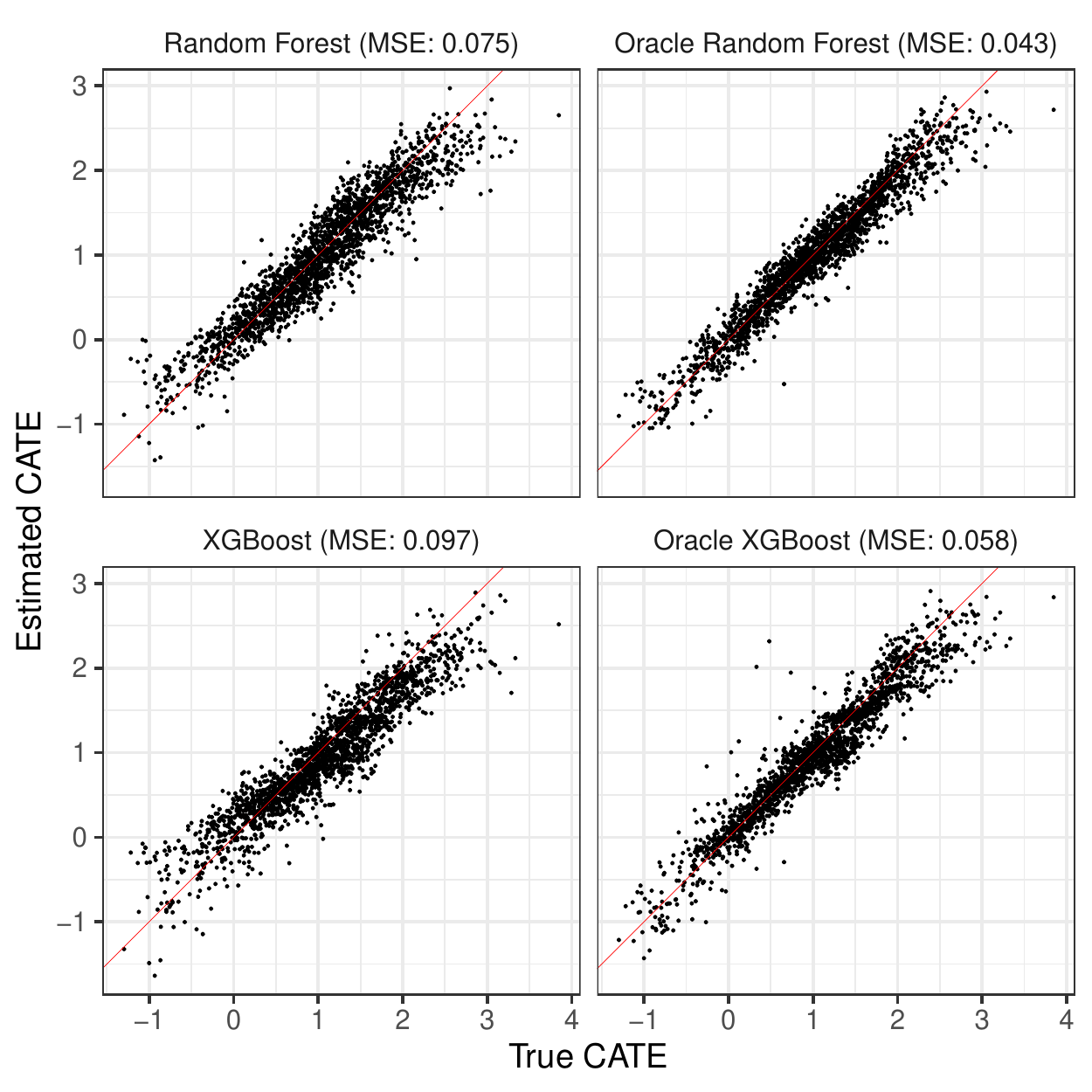}
\caption{Top: \emph{P-learner} implemented with honest regression forest (with default tuning parameters) \citep{athey2019generalized} fit on simulated training data ($n_{train}=4000$ and 5 covariates) where unconfoundedness fails. The left figure shows predictions on a test set ($n_{test}=2000$) vs true simulated CATEs when using $\hat h$ and $\hat q$ \citep{ghassami2022minimax} while the right-hand panel uses $h^*$ and $q^*$. Bottom: \emph{P-learner} implemented with \emph{XGBoost} \citep{chen2016xgboost} (with tuning parameters selected by cross-validation). Mean square error (MSE) is defined as $\frac{1}{n}\sum_{i=1}^{n_{test}}(\hat \tau(X_i) - \tau^*(X_i))^2$.
}
\label{fig:rfVSxgb}
\end{figure}

In the right-hand panel in Figure \ref{fig:cfVSlasso} we use \emph{glmnet} \citep{friedman2010regularization, Rcore} to fit a \emph{P-learner} using cross-validated Lasso \citep{tibshirani1996regression} on squared and pairwise interactions of a 7-degree natural spline-based expansion on $X$ where we use cross-fit estimates of nuisance components $h$ and $q$ using kernel estimation \citep{ghassami2022minimax}. This figure suggests the promise of the \emph{P-learner}, as the point estimates are not far off from an oracle learner on $h^*$ and $q^*$ (Figure \ref{fig:cfVSlasso} bottom panel). In Figure \ref{fig:rfVSxgb} we consider two different final stage learners for $\tau^*(\cdot)$: Random Forest \citep{breiman2001random}, fit using an honest\footnote{A honest regression forest use sample splitting to avoid estimation bias that arises from using the same observations to perform CART splitting as to form the leaf averages. For the example considered here, an honest regression forest performed better than a traditional random forest.} regression forest as implemented in \emph{grf} \citep{grf} using default tuning parameters, and boosting using \emph{XGBoost} \citep{chen2016xgboost} using tuning parameters selected by cross-validation. In this particular example \emph{XGBoost} has slightly more trouble adapting to the $\tau(\cdot)$ signal (choosing a good grid of tuning parameters may sometimes be challenging), though paints a slightly more realistic picture in the sense that the deviation from the oracle MSE might be large for some realizations. 

To conclude this section we caution that while the simulation example is intended to bear some resemblance to a real-world scenario where heterogeneity is present, but hidden beneath unmeasured confounding, it is but a toy example that has limited use in helping choose among different \emph{P-learners} in practice. A machine learning algorithm fit to minimize the empirical loss \eqref{eq:plearnereq} may do very well in minimizing test set error, regardless of whether treatment heterogeneity is actually present or not. Section \ref{sec:applicationRATE} describes our suggested approach for evaluating the \emph{practical} performance of a \emph{P-learner} by using the recently developed RATE metric by \citet{yadlowsky2021evaluating} which under considerable generality can be paired with the proximal doubly robust scores \eqref{eq:psi} to deliver a test set area-under-the-curve (AUC) measure of heterogeneity along with bootstrapped confidence intervals.\footnote{For ``real-world'' simulation-based approaches to validating estimators see \citet{athey2021using} and for further discussion on the limitations of benchmarking in the context of CATE estimation see \citet{curth2021really}.}

\section{Treatment Heterogeneity in the SUPPORT Study}\label{sec:application}
To illustrate the \emph{P-learner} in action we consider data from the Study to Understand Prognoses and Preferences for Outcomes and Risks of Treatments (``SUPPORT'', \citet{connors1996effectiveness}), used in a series of papers \cite{tchetgen2020introduction,cui2020semiparametric,qi2023proximal,ying2022proximal} as an example of the proximal causal inference framework. The treatment assignment in this study is a so-called right heart catheterization that was administered to certain patients when admitted to the intensive care unit. The outcome of interest is the number of days between admission, and death or censoring at 30 days. The SUPPORT study did not randomize treatment, however, an informative set of covariates were collected for each of the patients that were administered treatment (2184 patients) or control (3551 patients). \citet{connors1996effectiveness}'s original analysis concluded that heart catheterization was on average harmful to patient health. To account for latent confounding due to measurement error in the patient's physiological variables, \citet{tchetgen2020introduction} reanalyze the data by using proxy variables $Z=($pafi1, paco21$)$ and $W=($ph1, hema1$)$, a set of  physiological status measures, and find evidence (using parametric modeling) that the harmful effect is even larger (ATE = $-1.80$ days) than previously reported. To investigate the possibility of heterogeneity in this harmful effect we fit \emph{P-learners} using the covariates $($age, sex, cat1\_coma, cat2\_coma, dnr1, surv2md1, aps1$)$ also considered in \citet{ghassami2022minimax} for illustrating the non-parametric kernel estimation of $h$ and $q$. Among those variables, cat1\_coma, cat2\_coma, and dnr1 are indicator variables indicating coma and ``Do Not Resuscitate'' status, while surv2md1 and aps1 are estimates of 2-month survival and severity-of-disease score respectively. 

\begin{table}[ht!]
    \begin{center}
    \caption{Best linear projection (BLP) of the CATEs on covariates considered in \citet{ghassami2022minimax} for the SUPPORT data, as well as a doubly robust estimate of the average treatment effect (ATE), the BLP on a constant. All 5735 units are used to form cross-fit estimates of nuisance parameters $h$ and $q$, using the kernel method of \citet{ghassami2022minimax} with associated suggested hyperparameters for this study. $HC_3$ \citep{mackinnon1985some} standard errors are in parentheses.}
    \vspace{0.1in}
    \label{table:coefficients}
    \begin{tabular}{l c c}
    \hline
     & (BLP) & (ATE) \\
    \hline
    (Intercept) & $4.55 \; (2.41)$        & $-1.66 \; (0.27)^{***}$ \\
    age         & $-0.06 \; (0.02)^{***}$ &                         \\
    sex         & $-1.03 \; (0.54)$       &                         \\
    cat1\_coma  & $-0.28 \; (1.16)$       &                         \\
    cat2\_coma  & $3.08 \; (2.22)$        &                         \\
    dnr1        & $1.00 \; (0.87)$        &                         \\
    surv2md1    & $-3.93 \; (1.88)^{*}$   &                         \\
    aps1        & $0.01 \; (0.02)$        &                         \\
    \hline
    \multicolumn{3}{l}{\scriptsize{$^{***}p<0.001$; $^{**}p<0.01$; $^{*}p<0.05$}}
    \end{tabular}
    \end{center}
\end{table}

As a first step, we consider a linear CATE model, Table~\ref{table:coefficients} show estimates of the best linear projection 
\begin{equation}
\label{eq:BLP}
\{\beta^*_0, \, \beta^*\}  = \arg\min_{\beta_0,\beta}\, \EE{( \tau^*(X_i) - \beta_0 - X_i \beta)^2},
\end{equation}
using cross-fit estimates of $h$ and $q$ on all 5735 units to form the proximal scores \eqref{eq:psi}. The second column of Table \ref{table:coefficients} is simply the projection onto a constant and recovers an ATE in line with the estimates from \citet{ghassami2022minimax} and \citet{tchetgen2020introduction}.

Next, we consider the three \emph{P-learners} described in Section \ref{sec:illustration}. For the Lasso learner, we use the same spline-based featurization for the continuous covariates $X_i$ and just interactions for the binary $X_i$. For all learners, we fit $h$ and $q$ using the non-parametric kernel estimator of \citet{ghassami2022minimax} (using their hyperparameters suggested for this dataset).

\subsection{Assessing Treatment Heterogeneity with RATE}\label{sec:applicationRATE}
As pointed out in Section \ref{sec:illustration}, a fundamental challenge with evaluating CATE estimators on real-world data is the lack of ground truth for calculating traditional error metrics, as treatment effects are fundamentally unobservable. Recent developments in the statistics literature offer an attractive solution to this problem in the form of a family of metrics called rank-weighted average treatment effects (RATE). A RATE metric's fundamental ingredient is a calibration curve inspired by the ROC, called the Targeting Operator Characteristic (TOC), defined as:
\begin{align*}
&\text{TOC}(q) = \EE{Y_i(1) - Y_i(0) | \hat \tau(X_i) \geq F^{-1}_{\hat \tau(X_i)}(1 - q)}\\
& \ \ \ \ \ \ \ \ \ \ \ \ \ \ \ - \EE{Y_i(1) - Y_i(0)},
\end{align*}
and is a curve that ranks all observations on a test set according to $\hat \tau(X_i)$ estimated from a training set, and compares the ATE for the top $q$-th fraction of units prioritized by $\hat \tau(X_i)$ to the overall ATE. The RATE is the area under this curve and measures how well the CATE estimator stratifies the population in terms of units that benefit differently from treatment. If there is significant heterogeneity present, and the CATE estimator detects it, then the estimated RATE should be significantly different from zero. When selecting among CATE estimators, the one with the largest RATE metric is the best performing in the sense that it most successfully manages to stratify test set subjects according to different treatment benefits.

Inferential properties of the RATE extend to the proximal setting through the doubly robust proximal scores~\eqref{eq:psi} \citep[Theorem 4]{yadlowsky2021evaluating} which satisfy $\EE{\hat \Gamma_i | X_i} \approx  \EE{Y_i(1) - Y_i(0) | X_i}$ and motivates the following Algorithm 2 to evaluate a \emph{P-learner}.

\begin{alg}{(Evaluate P-learner)}
\end{alg}
\begin{description}
    \item[Step 1.] Randomly partition the data into a training and evaluation set.

    \item[Step 2.] Using Algorithm 1, learn a $\tau(\cdot)$ function on the training data.

    \item[Step 3.] Estimate cross-fit doubly robust scores \eqref{eq:psi} on the evaluation data and predict $\hat \tau(X_{evaluation})$ using $\tau(\cdot)$ learned in Step 2. Use these to compute the TOC and the area under the TOC.
\end{description}

\begin{figure}[ht!]
\centering
    \includegraphics[scale=0.4]{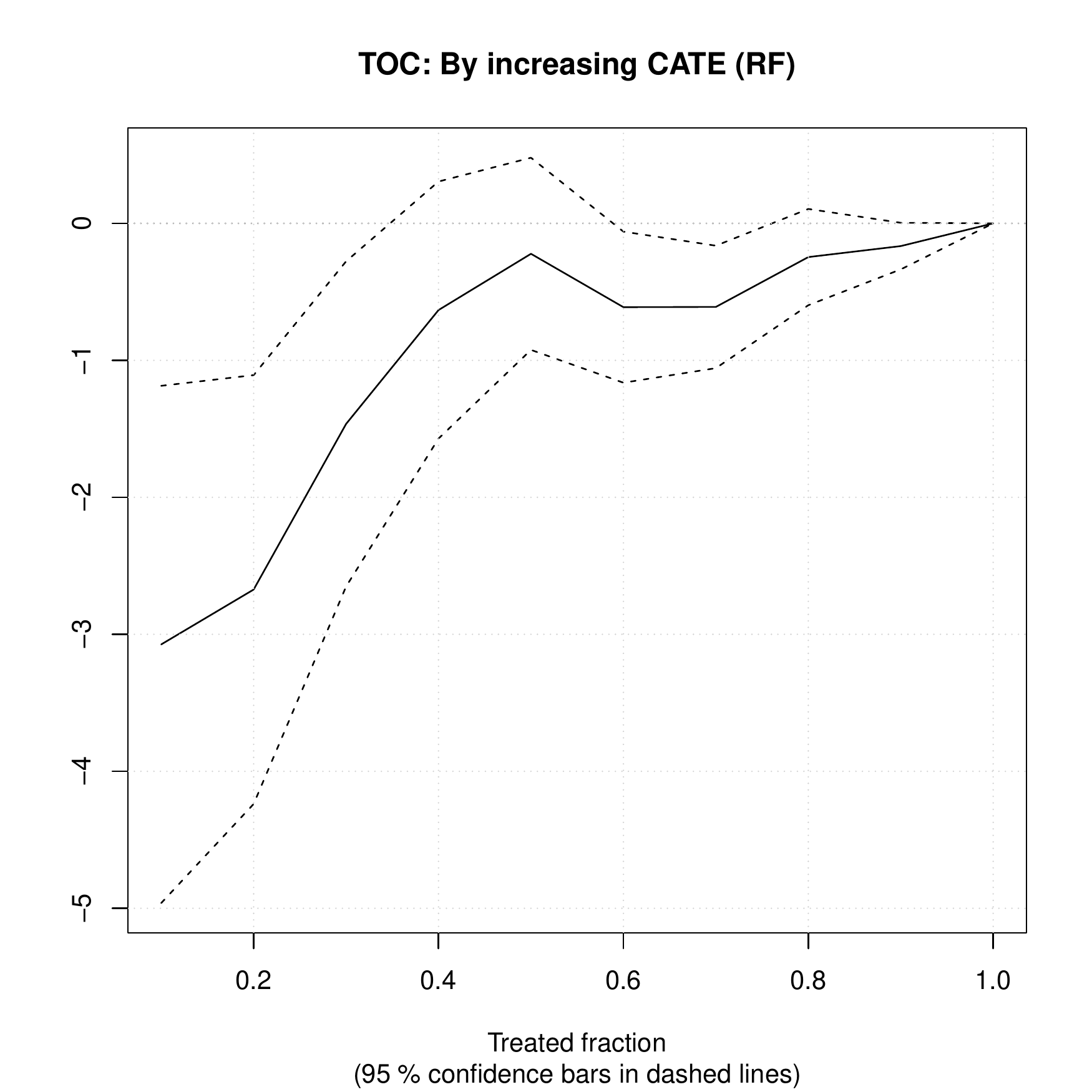}
\caption{The Targeting Operator Characteristic (TOC) for CATEs estimated with a random forest-based \emph{P-learner} fit on the SUPPORT data, using half of the dataset for CATE training, and the other half for evaluating the RATE. The area under the TOC curve (AUTOC) is $-1.07$ with a bootstrapped standard error of $0.31$.}
\label{fig:figTOC}
\end{figure}

Table \ref{table:RATEs} show estimated RATEs using three \emph{P-learners} and indicate that all procedures learn a CATE function that manages to stratify units on a test set\footnote{Since the ATE is negative we form the TOC by conditioning on the most negative CATEs first.}. The largest estimated RATE is obtained by the Random Forest-based \emph{P-learner} and Figure \ref{fig:figTOC} shows the corresponding TOC curve. The RATE is the area under this curve and has a point estimate of $-0.79$ with a bootstrapped standard error of $0.31$, and suggests there is heterogeneity in the response to right heart catheterization. For example, from Figure \ref{fig:figTOC}, the patients in the lowest $20$-th quantile of estimated CATEs, die 3 days sooner than on average when administered right heart catheterization, indicating a considerably more harmful effect for certain parts of the population than previously reported for the population as a whole.

\begin{table}[ht!]
    \begin{center}
    \caption{Estimated RATEs and bootstrapped standard errors using CATE estimates obtained by training \emph{P-learners} (selecting tuning-parameters with cross-validation) on a random half-sample of the SUPPORT study (with cross-fit kernel estimates of $h^*$ and $q^*$ with hyperparameters suggested in \citet{ghassami2022minimax}). Proximal doubly robust scores for evaluation are estimated on the held-out evaluation set. 
    }
    \vspace{0.1in}
    \label{table:RATEs}
    \begin{tabular}{l c c c}
    \hline
     & Lasso & Random Forest & XGBoost \\
    \hline
    AUTOC & $-0.79$ & $-1.07$ & $-0.72$ \\
    Std.err & $(0.39)$  & $(0.31)$ & $(0.34)$ \\
    \hline
    \end{tabular}
    \end{center}
\end{table}

\section{Discussion and Extensions}\label{sec:conclusion}
Detecting and measuring treatment effect heterogeneity is an important task across a wide range of domains, and in many observational applications, the investigator does not have access to an unconfounded treatment assignment. We have shown how powerful approaches to CATE estimation under unconfoundedness that rely on a Neyman-orthogonal loss can be extended to the seminal proximal causal learning framework of \citet{tchetgen2020introduction}.

A natural extension is to ask for the CATE conditional on proxy variables $Z_i$ and $W_i$: $\tau^*(x, z) = \EE{Y_i(1) - Y_i(0) | X_i = x, Z_i = z}$ and $\tau^*(x, w) = \EE{Y_i(1) - Y_i(0) | X_i = x, W_i = w}$. Given doubly robust scores for $\tau^*(x,z)$ and $\tau^*(x,w)$, these could in principle be used in place of \eqref{eq:psi} in the \emph{P-learner}. The derivation of these is an active research area (Tchetgen Tchetgen et al.)

Finally, a challenge with proximal causal learning is the estimation of the confounding bridge functions $h^*$ and $q^*$. \citet{kallus2021causal, ghassami2022minimax} make tremendously useful contributions in designing flexible regularized Gaussian kernel estimators for these components. Scaling this approach to large datasets, however, is challenging as kernel estimators typically require work on the order of $O(n^3)$. As pointed out by \citet{dikkala2020minimax}, some promising approximation schemes such as Nystr\"om's method can bring this down to the order of $O(n^2)$. \citet{kompa2022deep} take a neural network approach to bridge function estimation and avoids the reliance on kernel methods. Alternate approaches like this may suggest other avenues for scaling. One example could be the connection to the general minimax optimization problem \citep{dikkala2020minimax}: this is a type of adversarial estimation problem that is similar to the ones encountered and successfully solved at large scales with Generative Adversarial Networks \citep{goodfellow2014, arjovsky2017wasserstein}, and which have promising adaptions for certain statistical tasks  \citep{kaji2020adversarial, chernozhukov2020adversarial}.

\section*{Acknowledgements}
We are grateful to the ICML reviewers, as well as seminar participants at Stanford University, for their helpful and insightful comments. Yifan Cui gratefully acknowledges funding from the National Natural Science Foundation of China.

\bibliography{main}
\bibliographystyle{icml2023}

\newpage
\appendix
\onecolumn

\section{Proof of Proposition~\ref{prop}}

Let $P$ be a non-negative measure over the compact metric space $\cX \subset \mathbbm R^d$, and $K$ be a kernel with respect to $P$. Let $T_K: L_2(P) \rightarrow L_2(P)$ be defined as $T_K(f)(\cdot) = \EE{K(\cdot, X)f(X)}$. By Mercer’s theorem \citep{cucker2002mathematical}, there is an orthonormal basis of eigenfunctions $(\psi_j)_{j=1}^\infty$ of $T_K$ with corresponding eigenvalues $\{\sigma_j\}_{j=1}^\infty$ such that $K(x, y) =  \sum_{j=1}^\infty \sigma_j \psi_j(x) \psi_j (y)$. 

We consider the function $\phi: \cX \rightarrow l_2$ defined by $\phi(x) = \{\sqrt{ \sigma_j}\psi_j (x)\}_{j=1}^\infty$. Following \citet{mendelson2010regularization}, we define the RKHS $\cC$ to be the image of $l_2$, that is, for every $t \in l_2$, we define the corresponding element in $\cC$ by $f_t(x) = \langle \phi(x),t \rangle$, with the induced inner product $\langle f_s,f_t \rangle_{\cC}=\langle t,s\rangle$.

\begin{assu}\label{ass:kernel}
Without loss of generality, we assume $K(x, x) \leq 1$ for any $x \in \mathcal X$. We further assume that for $0 < p < 1$, $\sup_{j\geq 1} j^{1/p} \sigma_j = G_1 < \infty$ and  $\sup_j ||\psi_j||_{\infty} \leq G_2 <\infty $ for some constants $G_1$ and $G_2$.
\end{assu}

\begin{assu}\label{ass:operator}
We assume that $||T_K^\beta(\tau^*(\cdot))||_{\cC}<\infty$ for some $0 < \beta < 1/2$.
\end{assu}

\begin{assu}
For any $a$, we have that
$\EE{[\hat h (W,a,X)- h^*(W,a,X)]^2}=o(n^{-1/2})$, and $\EE{[\hat q (Z,a,X)- q^*(Z,a,X)]^2}=o(n^{-1/2})$.
\end{assu}

\begin{assu}
\label{assu:bound}
We assume that $|Y_i| \leq M$, $\sup_{w,a,x}|h(w,a,x)| \leq M$,  and
$\sup_{z,a,x}|q(z,a,x)| \leq M$.
\end{assu}

To give some intuition on the role of $p$ and $\beta$, $0 < \beta < 1/2$ essentially quantifies the amount of smoothing needed for $\tau^*$ to have finite $\cC$-norm, where 0 corresponds to $||\tau^*||_{\cC} < \inf$ and $1/2$ would mean $\tau^*$ is square-integrable. $p$ captures how much the $\tau$ function can oscillate through the eigenvalues of the corresponding RKHS. If $\beta, p \rightarrow 0$, then Theorem \ref{thm} yields the familiar result that $4$-th root nuisance rates are sufficient to yield $\sqrt{n}$ rates for a single target parameter. 

\begin{proof}

We have the following decomposition,
\begin{align*}
& \hat L(\tau)-\hat L(\tau_\alpha^*)-\tilde L(\tau)+\tilde L(\tau_\alpha^*)\\
= & \frac{2}{n} \sum_{i=1}^n 
- A_i[\hat h(W_i, A_i, X_i)- h^*(W_i, A_i, X_i)][\hat q(Z_i, A_i, X_i) - q^*(Z_i, A_i, X_i)](\tau_h^*(X_i)-\tau(X_i))\\
& + A_i [\hat q(Z_i, A_i, X_i) - q^*(Z_i, A_i, X_i)][Y_i - h^*(W_i, A_i, X_i)](\tau_\alpha^*(X_i)-\tau(X_i))\\
& +  [\hat h(W_i,1, X_i)-h^*(W_i,1, X_i)] [1- A_iq^*(Z_i, A_i, X_i)]](\tau_\alpha^*(X_i)-\tau(X_i)).
\end{align*}

By the Cauchy Schwarz inequality, the first term is bounded by 
\begin{align*}
\frac{2}{n} \sum_{i=1}^n 
& - A_i[\hat h(W_i, A_i, X_i)- h^*(W_i, A_i, X_i)] [\hat q(Z_i, A_i, X_i) - q^*(Z_i, A_i, X_i)](\tau_\alpha^*(X_i)-\tau(X_i))\\
\leq &  2 \sqrt{\frac{1}{n}\sum_{i=1}^n [\hat h(W_i, A_i, X_i)- h^*(W_i, A_i, X_i)]^2} \sqrt{\frac{1}{n}\sum_{i=1}^n[\hat q(Z_i, A_i, X_i) - q^*(Z_i, A_i, X_i)]^2}\times ||\tau_\alpha^*-\tau||_{\infty}\\
\leq & \alpha^{p} R(\tau,\alpha)^{\frac{1-p}{2}} O_p(n^{-1/2}),
\end{align*}
where the last inequality holds by the following fact
\begin{align*}
||\tau||_{\infty} \leq
||\tau||^p_{\cC}
||\tau||^{1-p}_{L_2(P)},
\end{align*}
as provided in Lemma 5.1 of \citet{mendelson2010regularization}.

Next, we consider the second term, and the third term can be bounded in a similar manner. 
For the $c$-th fold, we define
\begin{align*}
\eta_c(\tau,\alpha) = \frac{2}{n_c} \sum_{i=1}^{n_c}  A_i [\hat q^c(Z_i, A_i, X_i) - q^*(Z_i, A_i, X_i)][Y_i - h^*(W_i, A_i, X_i)](\tau_\alpha^*(X_i)-\tau(X_i)),
\end{align*}
where to ease notation, we use superscript $c$ to denote $(-c(i))$ and $n_c$ to denote the sample size of $c$-th fold. 
Note that to bound $\sup_{\tau\in \cC_{\alpha}}\{ \eta_c(\tau,\alpha)\}$, we essentially need to bound $\sup_{\tau\in \cC_{\alpha}}\{ \eta_c(\tau,\alpha): ||\tau-\tau_{\alpha}^*||_{L_2(P)} \leq L \}$.

We denote the samples that are not included in the $c$-th fold by $I^c$.
By cross-fitting, we have that
\begin{align*}
&\EE{\eta_c(\tau,\alpha)|I^c} \\
=&\frac{2}{n} \sum_{i=1}^n \EEE[A_i [\hat q^c(Z_i, A_i, X_i) - q^*(Z_i, A_i, X_i)][Y_i - h^*(W_i, A_i, X_i)](\tau_\alpha^*(X_i)-\tau(X_i))|I^c]\\
=&\frac{2}{n} \sum_{i=1}^n \EEE[\EEE[A_i [\hat q^c(Z_i, A_i, X_i) - q^*(Z_i, A_i, X_i)][Y_i - h^*(W_i, A_i, X_i)](\tau_\alpha^*(X_i)-\tau(X_i))|Z_i,A_i,X_i,I^c]I^c]\\
=&\frac{2}{n} \sum_{i=1}^n \EEE[A_i [\hat q^c(Z_i, A_i, X_i) - q^*(Z_i, A_i, X_i)] \EE{Y_i - h^*(W_i, A_i, X_i)|Z_i,A_i,X_i}(\tau_\alpha^*(X_i)-\tau(X_i))|I^c]\\
= & 0.
\end{align*}

Following Lemma~5 of \citet{nie2020quasi}, we further have that 
\begin{align*}
\frac{\EE{\sup_{\tau\in \cC_{\alpha}}\{ \eta_c(\tau,\alpha): ||\tau-\tau_{\alpha}^*||_{L_2(P)} \leq L \} | I^c}}{\EE{(A_i [\hat q^c(Z_i, A_i, X_i) - q^*(Z_i, A_i, X_i)])^2}^{1/2}} =O_p(\alpha^p L^{1-p} \frac{\log(n)}{n^{1/2}}),
\end{align*}
and therefore, we have
\begin{align*}
\EE{\sup_{\tau\in \cC_{\alpha}}\{ \eta_c(\tau,\alpha): ||\tau-\tau_{\alpha}^*||_{L_2(P)} \leq L \} | I^c} =O_p(\alpha^p L^{1-p} \frac{\log(n)}{n^{3/4}}).
\end{align*}

Note that for some constants $C_1,C_2>0$,
\begin{align*}
\sup_{\tau\in \cC_{\alpha}} \{ ||A_i [\hat q^c(\cdot) - q^*(\cdot)]
[Y_i - h^*(\cdot)](\tau_\alpha^*(\cdot)-\tau(\cdot))||_{\infty}: ||\tau-\tau_{\alpha}^*||_{L_2(P)} \leq L \} \leq C_1 \alpha^p L^{1-p},
\end{align*}
and 
\begin{align*}
&\sup_{\tau\in \cC_{\alpha}} \{ \EEE[ \{A_i [\hat q^c(Z_i,A_i,X_i) - q^*(Z_i,A_i,X_i)][Y_i - h^*(W_i,A_i,X_i)](\tau_\alpha^*(X_i)-\tau(X_i))\}^2 ]: ||\tau-\tau_{\alpha}^*||_{L_2(P)} \leq L \}
\\ &\leq 
C_2 \alpha^{2p} L^{2(1-p)}n^{-1/2}.
\end{align*}
Then by Talagrand’s concentration inequality, for any fixed $\alpha,L,\epsilon>0$,
\begin{align*}
\sup_{\tau\in \cC_{\alpha}}\{ \eta_c(\tau,\alpha)|I^c: ||\tau-\tau_{\alpha}^*||_{L_2(P)} \leq L\}
\leq 
 O \left( \alpha^p L^{1-p} \frac{\log(n)}{n^{3/4}} +
\frac{\alpha^p L^{1-p}}{n^{3/4}} \sqrt{\log(\frac{1}{\epsilon})}
+ \alpha^p L^{1-p}\frac{1}{n}\log(\frac{1}{\epsilon}) \right)
\end{align*}
holds with probability larger than $1-\epsilon$. 

By a similar construction of \citet{nie2020quasi}, we have the following bound for any $\alpha>1$ and $L\leq 4M$.
\begin{align*}
\sup_{\tau\in \cC_{\alpha}}\{ \eta_c(\tau,\alpha): ||\tau-\tau_{\alpha}^*||_{L_2(P)}\leq L\}
\leq 
O_P \Bigg( \alpha^p L^{1-p}\frac{\log(n)}{n^{3/4}} +  \frac{\alpha^p L^{1-p}}{n^{3/4}} \sqrt{\log(  \frac{\alpha n^{1/(1-p)}}{L^2}) } \\+ \alpha^p L^{1-p}\frac{1}{n} \log(  \frac{\alpha n^{1/(1-p)}}{L^2}) +  \frac{\alpha^p L^{1-p}}{n^{5/4}}  \Bigg).
\end{align*}

Finally, by combining three terms, we have that 
\begin{align*}
\hat R(\tau,\alpha) - \tilde R(\tau,\alpha)
\leq O_P \Bigg(  \alpha^{p}  n^{-1/2} R(\tau,\alpha)^{\frac{1-p}{2}} + \\
\alpha^p \frac{\log(n)}{n^{3/4}}  R(\tau,\alpha)^{\frac{1-p}{2}}+  \frac{\alpha^p}{n^{3/4}} \sqrt{\log(  \frac{\alpha n^{1/(1-p)}}{R(\tau,\alpha)}) } R(\tau,\alpha)^{\frac{1-p}{2}}\\+ \frac{\alpha^p}{n} \log(  \frac{\alpha n^{1/(1-p)}}{R(\tau,\alpha)}) R(\tau,\alpha)^{\frac{1-p}{2}}+  \frac{\alpha^p}{n^{5/4}}R(\tau,\alpha)^{\frac{1-p}{2}}  \Bigg),
\end{align*}
which completes the proof.
\end{proof}

\section{Illustrative Data Generating Process}\label{sec:DGP}
We adopt the setup from \citet{cui2020semiparametric}. Covariates $X$ are drawn from a normal distribution $N(0, 0.25 I_{d \times d})$ where $I_{d \times d}$ is the identity matrix and $d = 5$; $A$ is drawn from a Binomial distribution with success probability $(1 + \exp((0.125, 0.125, 0, 0, 0)^T X))^{-1}$; $Z, W, U$ are drawn from a multivariate normal:

$$ 
(Z, W, U) | A, X \sim MVN\left(
    \mu = \begin{bmatrix}
    0.25 + 0.25A + (0.25, 0.25, 0, 0, 0)^T X \\
    0.25 + 0.125A + (0.25, 0.25, 0, 0, 0)^T X \\
    0.25 + 0.25A + (0.25, 0.25, 0, 0, 0)^T X
    \end{bmatrix},
    \Sigma = \begin{bmatrix}
    1 & 0.25 & 0.5\\
    0.25 & 1 & 0.5\\
    0.5 & 0.5 & 1
    \end{bmatrix}
    \right);
$$
and $Y$ is drawn from a normal with distribution with $\sigma = 0.25$ and conditional mean

$$
\EE{Y | W, U, A, Z, X} = 2 + \tau(X)A + (0.25, 0.25, 0, 0, 0)^T X + 2\EE{W | U, X} + 2W,
$$

where $\EE{W | U, X} = 0.25 + (0.25, 0.25, 0, 0, 0)^T X + 0.5(U - 0.25 - (0.25, 0.25, 0, 0, 0)^T X)$ and the treatment effect $\tau^*(X) = \exp(X_{(1)}) - 3X_{(2)}$ is from \citet{shen2022optimal}.

\end{document}